\documentclass[runningheads]{llncs}
\usepackage{graphicx}
\usepackage{amsmath,amssymb} 
\usepackage{color}
\usepackage{array}
\usepackage{comment}
\usepackage{hyperref}
\graphicspath{{figures/}}
\usepackage{xcolor}
\definecolor{darkblue}{rgb}{0.3, 0.3, 0.7}
\newcommand{\high}[1]{{\color{darkblue}#1}}

\newcommand{\fig}[1]{Figure~\ref{fig:#1}}
\newcommand{\tab}[1]{Table~\ref{tab:#1}}

\usepackage{multirow}
\usepackage{wrapfig}

\usepackage{graphicx}
\usepackage{amsmath}
\usepackage{amssymb}
\usepackage{booktabs}
\usepackage{comment}
\usepackage{booktabs}

\usepackage{xspace}
  \newcommand{\latinphrase}[1]{\textit{#1}}  

\newcommand{\eg}{\latinphrase{e.g.}\xspace}

\begin{document}
\title{Using phase instead of optical flow\\ for action recognition}

\titlerunning{Using phase instead of optical flow for action recognition}
\authorrunning{O. Hommos, S.L. Pintea, P.S.M Mettes, J.C. van Gemert}
\author{
    Omar Hommos$^1$,
    Silvia L. Pintea$^1$,\\
    Pascal S.M. Mettes$^2$,
    Jan C. van Gemert$^1$
}
\institute{
    \small $^1$Computer Vision Lab, Delft University of Technology, Netherlands\\
    \small $^2$Intelligent Sensory Interactive Systems, University of Amsterdam, Netherlands\\
}

\maketitle
\begin{abstract}
Currently, the most common motion representation for action recognition is optical flow. 
Optical flow is based on particle tracking which adheres to a Lagrangian perspective on dynamics. 
In contrast to the Lagrangian perspective, the Eulerian model of dynamics does not track, but describes local changes. 
For video, an Eulerian phase-based motion representation, using complex steerable filters, has been successfully employed recently for motion magnification and video frame interpolation. 
Inspired by these previous works, here, we proposes learning Eulerian motion representations in a deep architecture for action recognition.
We learn filters in the complex domain in an end-to-end manner.
We design these complex filters to resemble complex Gabor filters, typically employed for phase-information extraction.
We propose a phase-information extraction module, based on these complex filters, that can be used in any network architecture for extracting Eulerian representations. 
We experimentally analyze the added value of Eulerian motion representations, as extracted by our proposed phase extraction module, 
and compare with existing motion representations based on optical flow, on the UCF101 dataset.   
\end{abstract}

\begin{keywords}
Motion representation, phase derivatives, Eulerian motion representation, action recognition. 
\end{keywords}

\section{Introduction}
\label{sec:intro}
\begin{figure}[t]
    \centering
    \small
	\begin{tabular}{c@{\hskip 0.1in}c@{\hskip 0.1in}c@{\hskip 0.1in}c}
        \includegraphics[width=0.23\textwidth]{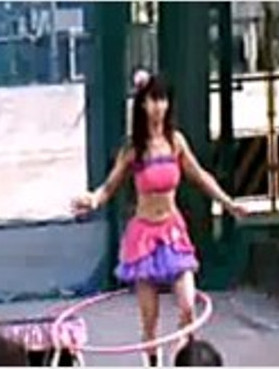} & 
        \includegraphics[width=0.23\textwidth]{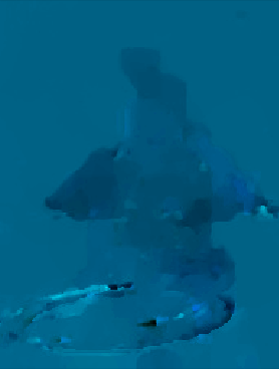} & 
        \includegraphics[width=0.23\textwidth]{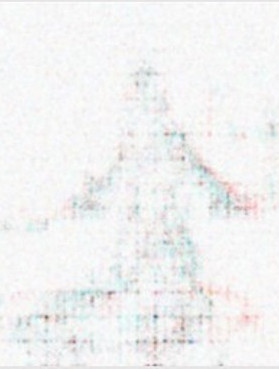} & 
        \includegraphics[width=0.23\textwidth]{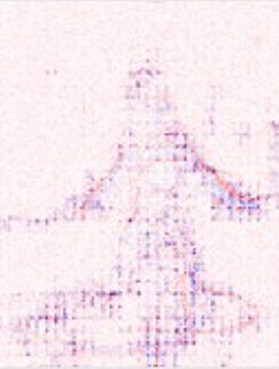} \\ [5px] 
        \includegraphics[width=0.23\textwidth]{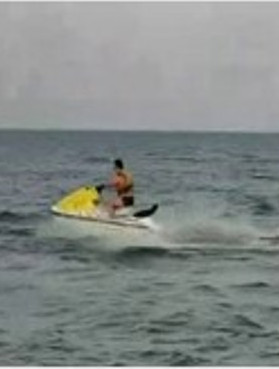} &
        \includegraphics[width=0.23\textwidth]{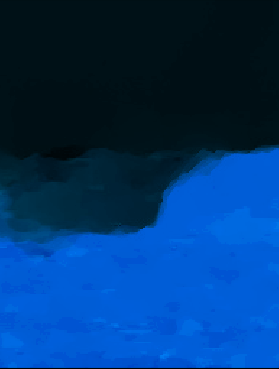} & 
        \includegraphics[width=0.23\textwidth]{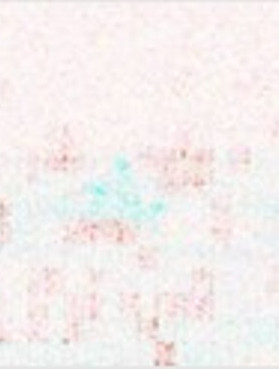} & 
        \includegraphics[width=0.23\textwidth]{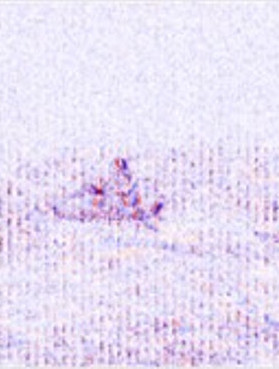} \\
        (a) Original input. & (b) Optical flow. & (c) RGB derivative. & (d) Phase derivative. \\
    \end{tabular}
    \caption{\textbf{Motion representations:} (a) Original input. (b) Optical flow visualized in HSV space, 
        where hue indicates the direction, and the saturation is the motion magnitude. 
        (c) Difference of RGB (dRGB), over time. 
        (d) Difference of phase (dPhase), over time (as described in section~\ref{sec:motion-description}). 
        For \emph{dRGB} and \emph{dPhase}, red indicates a positive value, while blue indicates a negative value.
        The optical flow fails to capture the motion of the waves around the boat. 
        The Eulerian representations better describe the motion at the boundaries of the objects.   
    }
    \label{fig:fig1}
\end{figure}

Recent advances in action classification rely on training Convolutional Neural Networks (ConvNets) on large video datasets \cite{quo-vadis,kinetics,two-stream,c3d}. Such ConvNets need to learn a suitable representation for motion, as it is important for discriminating similar actions that occur in a visually similar context: \eg \emph{Basketball Throw} and \emph{Basketball Dunk}~\cite{jain201515}.
Motion information is typically modeled by using optical flow as input to a separate network stream \cite{two-stream}, or by using 3$D$ convolutions \cite{c3d}, 
on stacks of input RGB frames and optical flow field stacks \cite{quo-vadis,fusion,actionflownet,laura,two-stream,rethinking}.
In this paper, inspired from the suggestions in~\cite{pintea2016making}, we zoom in on an alternative to using optical flow as a motion representation for deep action recognition in video: we investigate using phase in the complex domain instead of optical flow.

Optical flow follows the Lagrangian perspective on motion representations: tracking a pixel over time using its appearance, and the appearance information of neighboring pixels. 
Dissimilarly, an Eulerian motion representation focuses on the change in image information at a fixed spatial location, over time. 
Lagrangian methods require explicit point matching to obtain tracks, which is difficult on untextured surfaces, \eg water, hula hoops, or in the presence of occlusion. 
Eulerian motion representations, instead, do not need to compute explicit correspondences, yet they are sensitive to sudden large motions. 
\fig{fig1} depicts this difference between optical flow and Eulerian representations, defined as RGB and phase derivatives over time, on a few video examples. 

Previously, Eulerian representations based on RGB differences, have been considered in \cite{tsn}. 
However, they were only used as an input to the same network architecture employed for RGB inputs. 
The performance is highly dependent on the choice of network architecture \cite{quo-vadis}. 
Hence, architecture-level changes are necessary to make full use of the Eulerian motion information.
In this work, we propose a phase-extraction module composed of a complex convolutional layer followed by an arctangent function. 
The proposed module can be trained end-to-end and can be integrated in any existing network architecture.

This work brings forth: 
(i) the use of Eulerian motion representations for action recognition; 
(ii) learning phased information in an end-to-end manner by using  convolutional layers in the complex domain and complex activations; 
(iii) an empirical analysis of the advantages and failure cases for the phase-based motion representations, as well as a comparison with 
existing optical flow-based motion representations on the UCF101 dataset.

\section{Related work}
\label{sec:related_work}
\noindent \textbf{Learning action recognition.}
Top performing action recognition architectures use two-stream networks \cite{spatiotemporal-resnets,fusion,two-stream}.
When using a separate motion stream, the input is typically optical flow, and the stream uses popular architectures such as VGG-16 or ResNet, 
and inter-stream fusion \cite{spatiotemporal-resnets,fusion}. 
In this work, we adapt the motion stream of \cite{two-stream} architecture for our final network. 
However, our motion stream describes Eulerian motion.

A natural extension to video is the use of 3$D$ convolutions proposed in \cite{c3d}.
In \cite{quo-vadis}, the Inception architecture \cite{inception} enhanced with 3$D$ convolutions proved effective.
This architecture, coined \emph{I3D}, coupled with a new large action recognition dataset, Kinetics \cite{kinetics}, delivered state-of-the-art performance. 
In \cite{rethinking} 3$D$ convolutions using $k_t \times k \times k$ filters, are replaced with the more effective $1 \times k \times k$ followed by $k_t \times 1 \times 1$ filters, 
to reduce computational costs. 
In this work, we do not consider 3$D$ convolutions, but rather focus on learning Eulerian motion descriptions.

In \cite{varol2018long} long term convolutions based on optical flow are proposed for action recognition.
Recurrent Neural Networks (RNN) are also successfully used for modelling temporal information \cite{lstm1,lstm2}. 
Our proposed model can be used in combination with such architectures.
Where our proposed complex layer plays the feature extraction role. 

\bigskip
\noindent \textbf{Motion representations for action recognition.} 
To increase inference speed, in \cite{tsn} RGB differences are used instead of optical flow. 
It does not outperform optical flow, yet it obtains comparable performance while being $\times 25$ faster at inference-time. 
The use of motion vectors, that are similar to optical flow but capture only coarse-motion, is proposed in \cite{motion-vectors}.  
The motion vectors obtain a $\times 27$ speed improvement over two-stream networks \cite{two-stream}, while having comparable performance.
Unlike these methods, we propose learning Eulerian motion representation in an end-to-end framework for tackling action recognition where optical flow fails. 

A cascade of networks that learn to generate optical flow for the task of action recognition is used in \cite{actionflownet,laura}.
A similar motion representation is used in \cite{hidden-two-stream} before the classification network. 
Explicit use of optical flow as an input remains superior over other methods for action recognition. 
However, our aim here is to research if we can find a complementary motion representation to the optical flow.

\bigskip
\noindent \textbf{Phase-based methods.}
Video phase information has been successfully used before for tasks such as motion magnification and video frame interpolation.
Work such as \cite{fleet,gautama} reconstruct optical flow by extracting phase changes from image sequences and optimizing for velocity.
Measuring phase difference of a stereo image pair helps in estimating disparity at each pixel location for depth estimation \cite{disparity}.  
In \cite{kooij2016depth,pbmoma,zhang2017video} motion magnification of small movements was possible by measuring phase variations using a complex steerable pyramids over a sequence of images, 
and then magnifying motion in a reconstructed video.
In \cite{learn-pbmoma}, a ConvNet architecture was proposed to encode, manipulate, and then decode two subsequent input frames, to obtain an output frame with magnified motion. 
PhaseNet was proposed in \cite{phasenet}, which is a decoder network that receives the decomposition of two input frames, the result of applying steerable pyramid filters, 
and tries to predict the decomposition of the target frame. 
Similar to these works, and inspired by~\cite{pintea2016making} we adapt the idea of phase-based motion measurement, however dissimilar to previous works we propose to learn this end-to-end, through complex convolutions.

\section{Phase-Based Motion Description}
\label{sec:motion-description}
Fourier\rq s shift theorem states that a shift in time domain corresponds to a related linear phase shift in the frequency domain. 
Since phase variations directly correspond to change \cite{freeman1991design,learn-pbmoma,pbmoma}, they can serve as a viable representation of motion.
This does not only apply to the phase of the Fourier basis functions (sine waves), 
but also to the phase of other  representations in the complex domain such as complex-steerable pyramids \cite{pbmoma}.
Here, we build on the idea of representing local motion through the phase responses obtained by using a complete set of complex filters.
\subsection{Complex Filters for Motion Description}
Fleet and Jepson \cite{fleet} showed that the temporal evolution of contours of constant phase provides a good approximation to the motion field. 
In \cite{fleet,gautama} complex quadrature filters are used to extract the contours of constant phase. 
The temporal derivative of these responses is then employed to estimate object velocity in videos.    
Specifically, a set of complex Gabor quadrature filters are used for extracting phase information.
A complex quadrature Gabor filter, $H(x, y;  \lambda, \theta, \psi, \sigma, \gamma)$, is defined as:
\begin{alignat}{2}
H(x, y;  \lambda, \theta, \psi, \sigma, \gamma) 
    & = G(x, y;  \lambda, \theta, \high{\psi}, \sigma, \gamma) + i G(x, y;  \lambda, \theta, \high{\frac{\pi}{2} - \psi}, \sigma, \gamma)\\ 
    & = \exp \Bigg(-\frac{{x^\prime}^2 + \gamma ^2 {y^\prime}^2}{2 \sigma ^ 2}\ \Bigg) \times \exp \Bigg( i \big(2 \pi \frac{x^\prime}{\lambda} + \psi \big) \Bigg), 
\label{eq:cgabor}
\end{alignat}
where $G(\cdot)$ is a standard Gabor filter, $i=\sqrt{-1}$, $x^\prime = x \cos(\theta) + \text{y} \sin(\theta)$, and $y^\prime = -\text{x} \sin(\theta) + y \cos(\theta)$, 
$\lambda$ is the wavelength of the wave, $\theta$ represents its orientation, $\psi$ is its phase offset, $\sigma$ is the standard deviation of the Gaussian envelope, 
and $\gamma$ is the spatial aspect ratio of the Gaussian, used to control its ellipticity. 

In our case, we to not wish to precisely estimate the velocity of objects over time in the video, but rather describe the motion. 
Therefore, we relax the need of using quadrature filters for finding contours of constant phase and instead, we opt for the more simple perpendicular complex filters.

\bigskip\noindent\textbf{Perpendicular filters.} 
Learning complex quadrature filters using convolutional networks is difficult as it requires regularizing to ensure that the phase shift between the real and imaginary filters is $\pi/2$.
Here we opt for using perpendicular filters.  
Combining information from two perpendicular orientations gives a more complete, but less-orientation sensitive, response. 

To ensure that the learned filters are perpendicular, we fix the real filters, so they do not receive gradients during the training.
We only update the imaginary filters during training and reinitialize the real filters as a $\pi/2$ rotated version of the imaginary filters. 
This choice avoids numeric problems when extracting the phase information as $\text{atan}(\frac{\mathbf{x}_i}{\mathbf{x}_r})$, 
where $\mathbf{x}_r$ and $\mathbf{x}_i$ are the real and imaginary responses, respectively. 

\begin{figure}[t]
    \centering	
    \begin{tabular}{c@{\hskip 0.2in}c@{\hskip 0.2in}c}
           \includegraphics[width=0.128\textwidth]{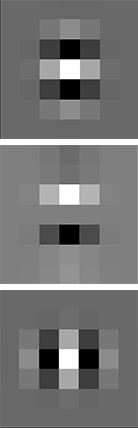} &
           \includegraphics[width=0.3\textwidth]{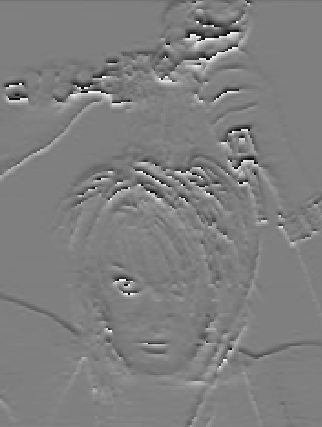} &
           \includegraphics[width=0.3\textwidth]{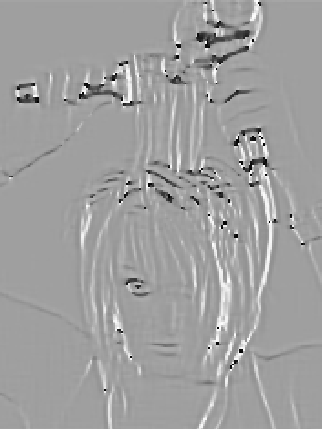} \\
            (a) Gabor filters. & (b) Quadrature responses. & (c) Perpendicular responses.\\
    \end{tabular}
    \caption{\textbf{Quadrauture versus perpendicular filters.} 
    (a) From top to bottom: the real and imaginary parts of a complex Gabor, followed by rotated real Gabor. 
    Top and middle form the complex quadrature pair, while top and bottom form a complex perpendicular pair. 
    (b) Phase responses to a complex quadrature Gabor. 
    (c) Phase response to a complex perpendicular filter. 
    In (b) only vertical orientations are highlighted, while both horizontal and vertical orientations are highlighted in (c), providing a more detailed description of the phase contours.}
    \label{fig:quadvspo}
\end{figure}

We find the perpendicular filters to be sufficient for describing motion contours in phase domain, and we validate this in our experiments.
\fig{quadvspo} depicts the difference between the responses of quadrature complex filters and perpendicular complex filters.

\bigskip\noindent\textbf{Sinusoidal Gabor regularization.}
To encourage the learned filters to resemble Gabor filters, we propose a regularization. 
We only train the imaginary part of our complex filters.
Since the imaginary Gabor filter is a sine multiplied by a Gaussian, 
we define the imaginary part of our complex filters as a multiplication between a filter initialized randomly and a non-trainable Gaussian kernel.    
We subsequently, regularize the trainable part of the filter to correspond to a sine.

In the Fourier domain a sine corresponds a single point, ignoring the domain symmetry. 
Thus, we minimize the $L_2$ distance from each point of the filter in the Fourier domain to the center of mass of the filter in the Fourier domain: 
\begin{alignat}{1}
    R(\mathbf{w}) = \sum_{i=1}^n \lVert \mathbf{w}_i - \text{CoM}(\mathbf{w}) \rVert_2,
\end{alignat}
where $n$ is the dimensionality of $\mathbf{w}$, $\mathbf{w}$ are the responses of the imaginary filters passed through a Real-FFT (Real Fast Fourier Transform), 
and $\text{CoM}(\cdot)$ computes the center of mass.

\begin{figure}[t]
    \centering \includegraphics[width=\textwidth]{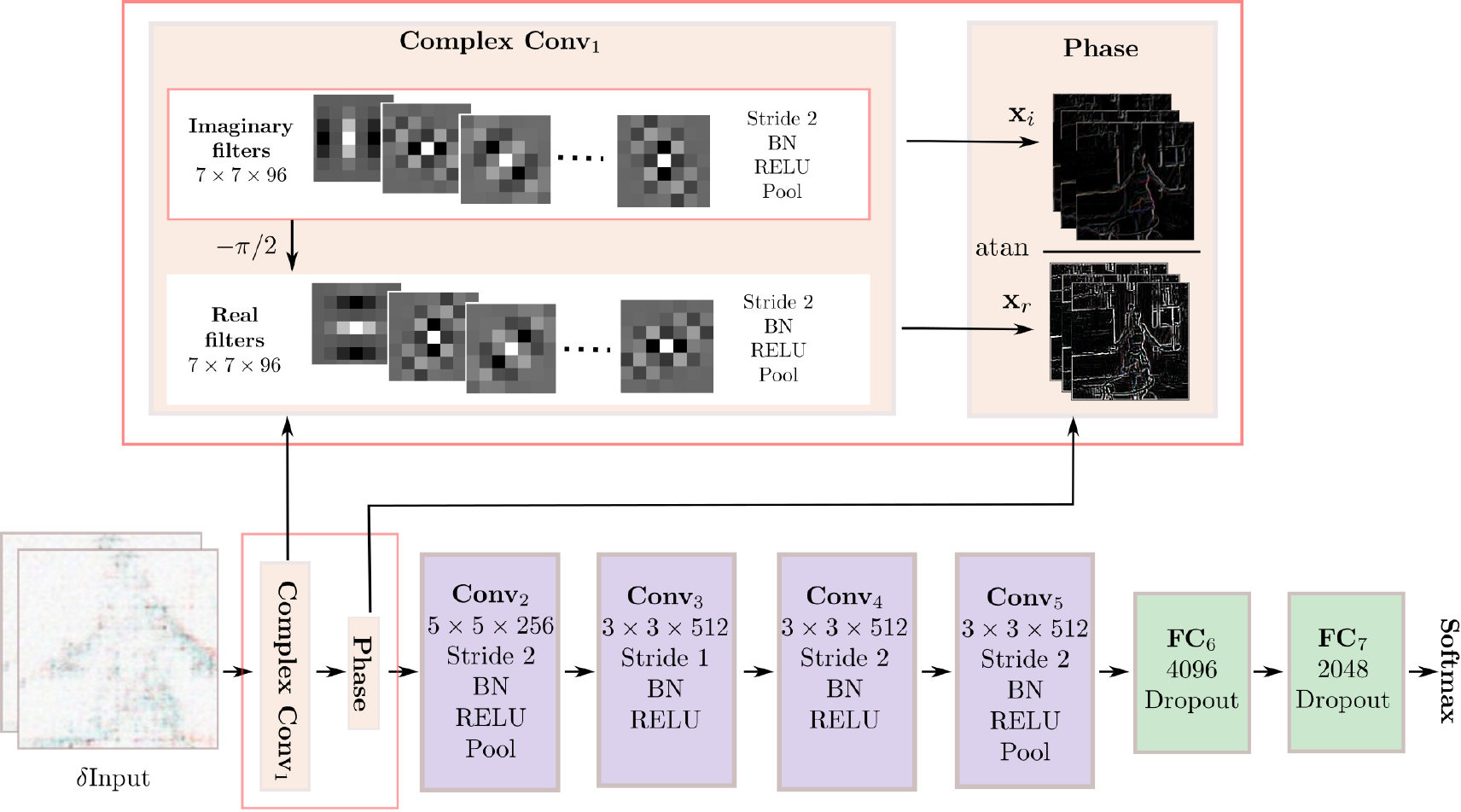}
    \caption{\textbf{Network Architecture.} 
        We input to our network temporal image derivatives. 
        From the input information, we learn perpendicular complex Gabor-like filters, in our proposed complex layer. 
        We use the responses of these filters to compute the phase.
        This information is subsequently send to the following layers in the network.
    }
    \label{fig:arch}
\end{figure}
\subsection{Learning Phase-Based Descriptions}
\label{sec:learning-descriptions}
We adapt a standard ConvNet to learn phase, by learning complex Gabor-like filters.
The trainable part of the filters is first initialized randomly.
Only the imaginary part of the filters is trained, and regularized with the proposed Gabor regularization.  
The real part of the complex filter is a $\pi/2$ rotated version of the imaginary filters.  
We use these learned filters to extract phase information, and we pass this information to the following layers for action recognition.  

Given that the temporal gradient of the phase is the one encoding the motion \cite{fleet}, we need to estimate temporal derivatives of the phase in our network.
However, differentiating the responses of a convolution is identical to differentiating one of the functions and 
then performing the convolution:
\begin{equation}
    \frac{\partial}{\partial t}(f * g) = \frac{\partial f}{\partial t} * g,
\end{equation}
\noindent
provided that two conditions hold: both functions $f$ and $g$ must be absolutely integrable, and $f$ must have an absolutely integrable ($L^1$) weak derivative \cite{bracewellconvolution}. 
Given this property, we input temporal image derivatives into our network, to estimated temporal derivatives of phase in our proposed module.

\fig{arch} displays our proposed network architecture: we input image derivatives, and from these we learn perpendicular Gabor-like complex filters.
We apply the complex non-linearity, $\mathbb{C}$ReLU, proposed in \cite{deepcomplex} after our complex convolutional layer.
$\mathbb{C}$ReLU effectively applies ReLU separately on the real and imaginary feature maps.
We also use standard BN (Batch Normalization).
We subsequently, estimate the phase as the arctangent of the responses of these filters, and we send this information to the following layers.

\section{Experiments}
\subsection{Experimental setup}
We use the network architecture displayed in \fig{arch}.
This is a replica of VGG-M, corresponding to one stream in \cite{two-stream}, but in which the first layer is replaced with our complex layer. 
For clarity we will refer to it as: \emph{PhaseStream}.
All experiments are performed on UCF101 \cite{ucf}, containing 101 action classes, with an average of 180 frames/video. 
We follow the standard training/testing data splitting. 
For the \textbf{Exp 1}, where we analyze design choices, we evaluate using only one standard data split. 
While, for \textbf{Exp 2}, we evaluate using the three standard data splits on UCF101.
For all experiments, we use momentum SGD as an optimizer with momentum of 0.9. 
Videos are uniformly sampled from all classes to create a batch of 256.
The dropout ratio is set to 0.9 and the learning rate is set to 0.01 and reduced by a factor of 10 at iterations 45000 and 75000.
We train for 100,000 iterations. 
Data is augmented with random crops and flips. 
\begin{figure}[t]
    \centering
    \begin{tabular}{c@{\hskip 0.3in}c}
           \includegraphics[width=0.45\textwidth]{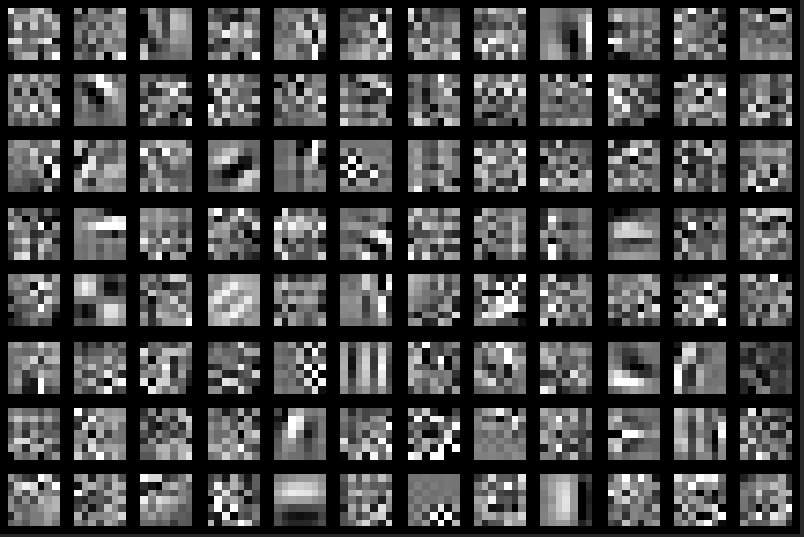} &
           \includegraphics[width=0.45\textwidth]{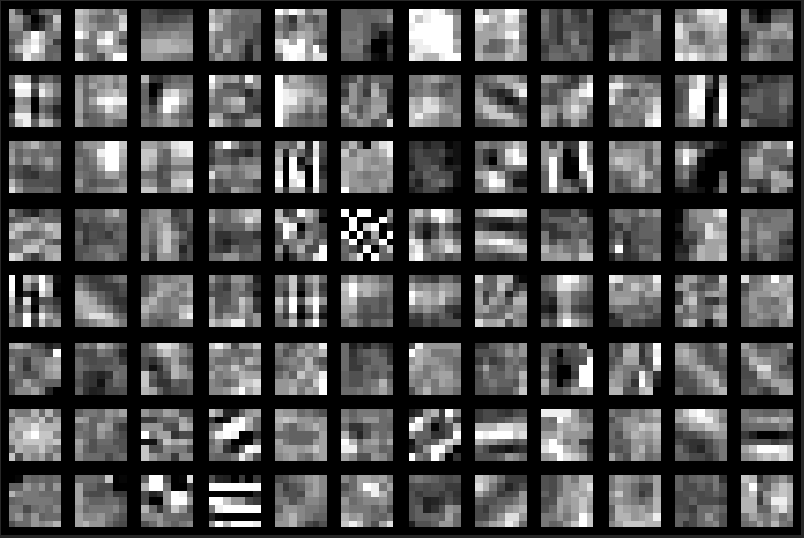} \\
            (i) Initialized filters. & (ii) Regularized complex filters.
    \end{tabular} 
    \caption{\textbf{Exp 1.(a):}
        The effect of the sinusoidal Gabor regularization: (i) randomly initialized filters; (ii) trained filters with the sinusoidal Gabor regularization. 
        The Gabor regularization compels the learned filters to be more similar to Gabor filters. 
    }
    \label{fig:filters}
\end{figure}
\begin{table}[t]
    \centering
    \begin{tabular}{l@{\hskip 0.2in}r@{\hskip 0.2in}r@{\hskip 0.2in}r}
        \toprule
        Filter type         & No. of  filters   & Training acc.    & Testing acc. \\ \midrule
        Quadrature Gabor    & 24                & $\sim$90 \%               & 60.5 \% \\
        Perpendicular Gabor & 24                & $\sim$70 \%               & 64.8 \% \\ 
        Perpendicular Gabor & 96                & $\sim$80 \%               & 71.6 \% \\
        \bottomrule
    \end{tabular}
    \caption{\textbf{Exp 1.(a):} Accuracy on UCF101 when using the proposed perpendicular complex filters when compared to quadrature Gabor filters.
        The network receives as input grayscale image derivatives over time, \emph{dGray}. 
        The perpendicular filters tend to generalize better than the quadrature filters.
    }
    \label{tab:benchmark}
\end{table}

\subsection{Exp 1: Importance of Eulerian information}
\noindent\textbf{Exp 1.(a): The use of perpendicular versus quadrature filters.}
To quantify the quality of learned perpendicular filters, we initialize our proposed complex layer with rotated Gabor filters. 
These filters are fixed throughout training. 
The results serve as a benchmark for the learning process. 
To compare quadrature filters with the perpendicular ones, our complex layer is initialized with quadrature Gabor filters, also fixed throughout training. 

\tab{benchmark} shows the results of the experiment. 
We consider two settings for the filter banks: one consisting of 24 filters, similar to the one in \cite{fleet}; and the other consisting of 96 filters, 
covering 12 logarithmically spaced frequencies between 0.2 and 5 Hz, over the same 8 directions $\theta=\pi/8 \times \{0, 1, .., 7\}$.
From the experimental analysis we conclude that the perpendicular filters help the network generalize better to unseen data.

\fig{filters}.(i) shows the initial set of complex filters, with random initialization. 
While \fig{filters}.(ii) shows the learned filters in our complex layer with the incorporated sinusoidal Gabor regularization described in section~\ref{sec:learning-descriptions}.
The regularization is effective in encouraging the learned complex filters to resemble Gabor filters.

\begin{table}[t]
    \centering
    \begin{tabular}{c@{\hskip 0.3in}c}
        \begin{tabular}{l@{\hskip 0.1in}c@{\hskip 0.1in}c}
            \toprule
                        & VGG-M \cite{two-stream} & PhaseStream \\ 
            Input       &                         & (our) \\ \midrule
            RGB         & \textbf{52.3} \% & 51.3 \% \\  
            OF          & 67.7 \% & N\slash A \\ [5px]
            dRGB        & 45.5 \% & \textbf{48.8} \% \\
            dGray       & 74.3 \% & \textbf{74.4} \% \\     
            dPhase      & 65.4 \% & \textbf{70.1} \% \\  
            \bottomrule
        \end{tabular} &
        \begin{tabular}{l@{\hskip 0.1in}c@{\hskip 0.1in}c}
            \toprule
                        & VGG-M \cite{two-stream} & PhaseStream \\ 
            Input       &                         & (our) \\ \midrule
            5$\times$OF         & 80.4 \% & N\slash A \\ [5px]
            5$\times$dGray      & 68.7 \% & \textbf{75.3} \% \\
            5$\times$dPhase     & \textbf{70.8} \% & 68.2 \%\\ 
            \bottomrule
        \end{tabular} \\ 
        (i) Different inputs. & (ii) Stacked inputs.\\[10px]
    \end{tabular}
    \caption{\textbf{Exp 1.(b):} Accuracy (\%) on UCF101 for different network inputs for our proposed \emph{PhaseStream} compared to 
        the VGG-M \cite{two-stream}.
        (i) We compare Eulerian representation: \emph{dGray} -- derivative of grayscale inputs over time, \emph{dRGB} -- derivative of RGB frames, 
        \emph{dPhase} -- derivative of phase images, with \emph{OF} -- optical flow, and RGB.
        The \emph{dPhase} and \emph{dGray} are stronger than using OF as input. 
        (ii) We consider also stacked inputs: \emph{5$\times$OF} -- OF stacked over 5 frames, \emph{5$\times$dGray} -- 5 grayscale derivatives stacked,
        and \emph{5$\times$dPhase} -- 5 phase derivatives stacked. 
        When stacking OF, \emph{5$\times$OF}, there is a substantial gain over stacked Eulerian inputs.
        Overall, using the proposed \emph{PhaseStream} is beneficial for all inputs except for stacked \emph{dPhase}, \emph{5$\times$dPhase}.
        We highlight in bold the network architecture with the highest accuracy. 
    }
    \label{tab:inputs}
\end{table}

\begin{table}[t]
    \centering
    \begin{tabular}{l@{\hskip 0.2in}l@{\hskip 0.3in}l@{\hskip 0.2in}l}
    \toprule
    VGG-M \cite{two-stream}   & $\Delta \%$ & PhaseStream (our)     & $\Delta \%$    \\ \cmidrule(l){1-2} \cmidrule(l){3-4} 
    Archery                   & 19.5 \%     & FloorGymnastics       & 19.4\% \\
    JumpingJack               & 15.8 \%     & TennisSwing           & 16.3 \% \\
    Rowing                    & 13.9 \%     & BoxingPunchingBag     & 14.3 \% \\
    CricketShot               & 12.2 \%     & WalkingWithDog        & 13.9 \% \\
    Skijet                    & 10.7 \%     & GolfSwing             & 12.8 \% \\
    BlowDryHair               & 10.5 \%     & MoppingFloor          & 11.8 \% \\
    PlayingFlute              & 10.4 \%     & HighJump              & 10.8 \% \\
    ApplyLipstick             & 9.4  \%     & UnevenBars            & 10.7 \% \\
    PlayingCello              & 9.1  \%     & ShavingBeard          & 9.3  \% \\
    HulaHoop                  & 8.8  \%     & HandstandWalking      & 8.8  \% \\ \bottomrule
    \end{tabular}
    \caption{
        \textbf{Exp 1.(b):} The relative improvements in accuracy on UCF101 between VGG-M \cite{two-stream} and our \emph{PhaseStream} with \emph{dGray} as input. 
            We show the top 10 classes for each one of the architectures. 
            The networks learn complementary information. 
    }
    \label{tab:analysis}
\end{table}
\bigskip\noindent\textbf{Exp 1.(b): The importance of the input.}
\tab{inputs} shows the performance of different inputs on two network architectures: the VGG-M \cite{two-stream} and our variant of VGG-M in which we replace
the first convolution with a complex convolution. 
We refer to it as \emph{PhaseStream}.
In \tab{inputs}.(i) we consider Eulerian inputs: \emph{dRGB} -- derivative of RGB frames obtained by temporally subtracting 2 consecutive frames, 
\emph{dPhase} -- derivative of phase frames, \emph{dGray} -- derivative of grayscale frames; 
and non-Eulerian inputs: \emph{RGB} and OF (Optical Flow).
In \tab{inputs}.(ii) we consider stacked variants of the inputs: \emph{5$\times$OF} where we stack 5 consecutive OF inputs, \emph{5$\times$dGray} and \emph{5$\times$dPhase}.
The Eulerian inputs perform better than the non-Eulerian ones on the VGG-M. 
However, when stacking the inputs, OF outperforms the rest.
Our \emph{PhaseStream} obtains improved performance for all inputs except for \emph{5$\times$dPhase}. 
We do not evaluate our \emph{PhaseStream} network on OF inputs, as computing complex responses over OF does not seem theoretically informative.     
The stacked Eulerian representations do not perform well due to the large or fast motion, which may result in combining 
different motion patterns of different object or adding noise into the motion representation.   

\begin{figure}[h!]
   \scriptsize
   \centering
        \begin{tabular}{c@{\hskip 0.1in}c@{\hskip 0.1in}c@{\hskip 0.1in}c}
        \includegraphics[width=0.23\textwidth]{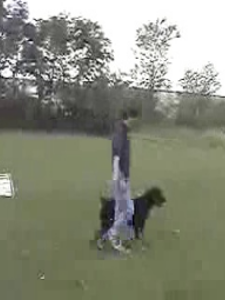} &
        \includegraphics[width=0.23\textwidth]{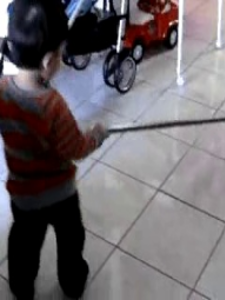} &
        \includegraphics[width=0.23\textwidth]{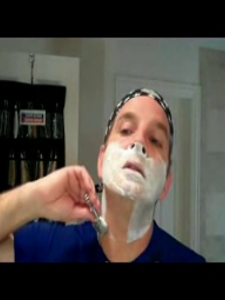} &
        \includegraphics[width=0.23\textwidth]{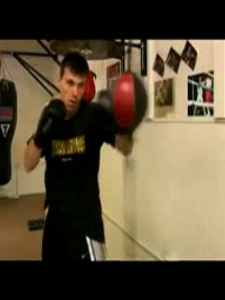} \\[5px]
        \includegraphics[width=0.23\textwidth]{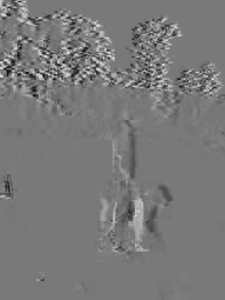} &
        \includegraphics[width=0.23\textwidth]{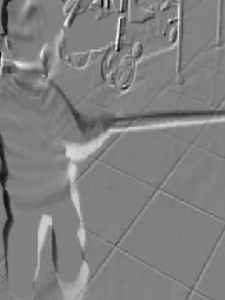} &
        \includegraphics[width=0.23\textwidth]{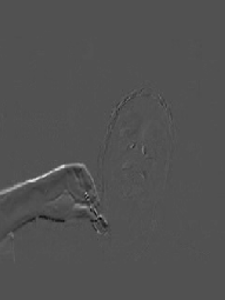} &
        \includegraphics[width=0.23\textwidth]{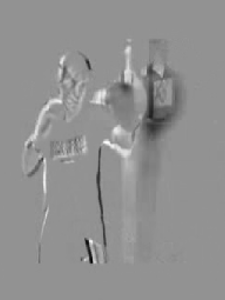} \\ [5px]
        \includegraphics[width=0.23\textwidth]{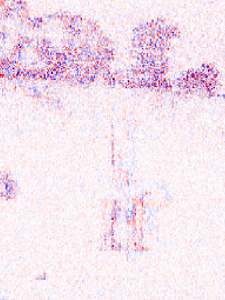} &
        \includegraphics[width=0.23\textwidth]{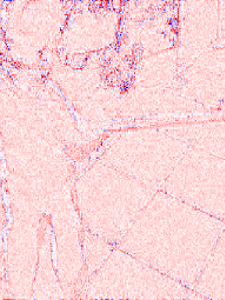} &
        \includegraphics[width=0.23\textwidth]{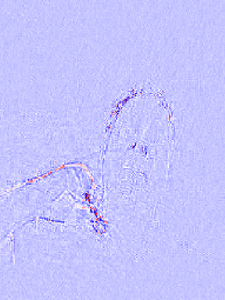} &
        \includegraphics[width=0.23\textwidth]{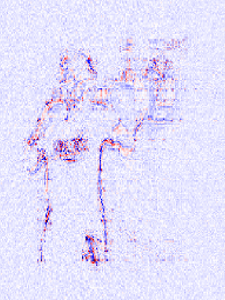} \\
        (i) WalkingWithDog. & (ii) MoppingFloor. & (iii) ShavingBeard. & (iv) BoxingPunchingBag. \\ 
        \end{tabular}
    \caption{
        \textbf{Exp 1.(b):} Examples of video frames from the classes \emph{BoxingPunchingBag}, \emph{WalkingWithDog}, \emph{MopppingFloor}, \emph{ShavingBeard},
        where the \emph{PhaseStream} performs better than the VGG-M \cite{two-stream}.
        We also display the temporal derivatives of grayscale inputs on the second row, and the temporal derivative of phase for these video frames, on the last row. 
        These classes are characterized by repetitive motion patterns. 
    }
    \label{fig:expl}
\end{figure}
\tab{analysis} shows the relative improvements in accuracy on the UCF101 dataset for VGG-M \cite{two-stream} and our \emph{PhaseStream} with grayscale frame derivatives, \emph{dGray}, as input. 
We show the top 10 classes with the largest improvements in accuracy. 
VGG-M performs better on action categories involving subtle motion, but which are more visual -- containing a specific visual object such as: \emph{Archery}, \emph{ApplyLipstick}, or \emph{PlayingFlute}.
The \emph{PhaseStream} obtains larger improvements over VGG-M from repetitive activities such as: \emph{BoxingPunchingBag}, \emph{WalkingWithDog}, \emph{MopppingFloor}, \emph{ShavingBeard}.
Examples of video frames from these categories are displayed in \fig{expl}.
We show the network inputs, \emph{dGray} inputs, on the second row.
On the last row we show the associated temporal derivatives of phase information, 
which is the type of information we would expect the network to rely on.

\bigskip\noindent\textbf{Exp 1.(c): Robustness of motion information.}
To quantify the robustness of motion information, we shuffle the input frames before calculating the temporal frame derivatives and OF, to be input to the network. 
This step effectively removes the temporal structure of the original video \cite{laura,rethinking}. 
We analyze how our proposed \emph{PhaseStream} performs when compared with the VGG-M \cite{two-stream} and TSN \cite{tsn}. 
For our \emph{PhaseStream} as well as for VGG-M, we use as input grayscale temporal image derivatives, \emph{dGray}.

\tab{shuffle} shows the test accuracies on UCF101, when feeding the networks standard inputs as well as temporally shuffled inputs. 
Optical flow numbers are taken from \cite{laura} for the TSN architecture \cite{tsn}. 
We also show the relative drop in performance, in percentages, caused by the loss of temporal ordering. 
The TSN performs the best in terms of absolute accuracy scores. 
When looking at the relative scores, the VGG-M has a slightly larger relative loss of performance, while TSN and our proposed \emph{PhaseStream} suffer a similar relative loss in accuracy. 
\begin{table}[t]
   \centering
        \begin{tabular}{l@{\hskip 0.2in}c@{\hskip 0.2in}c@{\hskip 0.2in}c}
        \toprule
        Architecture            & VGG-M \cite{two-stream}   & TSN \cite{tsn}    & PhaseStream (our) \\ 
        Input                   & dGray                     & 5$\times$OF       & dGray \\ \midrule
        Standard accuracy       & 74.3 \%                    & 86.9 \%            & 74.4 \% \\ 
        Shuffled accuracy       & 49.4 \%                    & 59.6 \%            & 51.4 \% \\ \midrule
        Relative change (\%)    & 33.5 \%                    & 31.4 \%            & 30.1 \% \\ 
        \bottomrule
    \end{tabular}
    \caption{
            \textbf{Exp 1.(c):} Test accuracies on UCF101 when training our \emph{PhaseStream}, VGG-M \cite{two-stream} and TSN \cite{tsn} 
            on standard inputs, as well as on temporally shuffled inputs. 
            We also show the relative drop in performance, in percentages. 
            The VGG-M suffers a slightly larger relative drop in performance, while TSN and our proposed \emph{PhaseStream} suffer a comparable drop in performance,
            when the temporal ordering is lost. 
    }
    \label{tab:shuffle}
\end{table}

\subsection{Exp 2: Comparison with existing work.}
\tab{comparison} shows the action recognition accuracy on the UCF101 dataset for a number of popular action recognition models: Two-Stream \cite{two-stream}, 
Two-Stream ResNets \cite{spatiotemporal-resnets}, TSN \cite{tsn}, Motion Vectors \cite{motion-vectors}, 
ActionFlow \cite{actionflownet} and our \emph{PhaseStream}.
We train our \emph{PhaseStream} on a stack of five differences of grayscale inputs, $5\times dGray$. 
Given that we focus on motion representations, we show the performance on the motion (temporal) stream only, for all the considered architectures.

TSN \cite{tsn} achieves the best performance as it relies on an ensemble of 3 two-stream networks, and provides several architecture-level improvements over them.
When it comes to Two-stream network \cite{two-stream}, Two-stream ResNets \cite{spatiotemporal-resnets}, and Motion Vectors \cite{motion-vectors}, 
the temporal input is a stack of 10 temporal representations. 
We use only a stack of 5 motion representations, as we did not see a great improvement from temporally stacking the Eulerian representations. 
Our proposed method on the temporal stream, obtains superior performance to ActionFlow \cite{actionflownet}, 
while having slightly lower performance than Motion Vectors \cite{motion-vectors}, and Two-stream ResNets \cite{spatiotemporal-resnets}.
These results validate that there is gain to be obtained from using Eulerian motion representations for action recognition.
\begin{table}[t]
    \centering
        \begin{tabular}{l@{\hskip 0.2in}c}
        \toprule
        Network                                                         & Motion (Temporal) Stream \\ \midrule 
        TSN \cite{tsn}                                                  & 83.8 \%   \\ 
        Two-Stream \cite{two-stream}                                    & 81.2 \%   \\
        Two-Stream ResNets \cite{spatiotemporal-resnets}                & 79.1 \%   \\
        Motion Vectors \cite{motion-vectors}                            & 79.3 \%   \\ 
        ActionFlow \cite{actionflownet}                                 & 70.0 \%   \\ [5px]
        PhaseStream (our) on \emph{$5\times dGray$}                     & 76.4 \%   \\ 
        \bottomrule
        \end{tabular}
    \caption{
        \textbf{Exp 2:} Accuracy (\%) on UCF101 comparing our proposed \emph{PhaseStream} trained on $5\times dGray$, 
        with the Two-stream \cite{two-stream}, Two-stream ResNet \cite{spatiotemporal-resnets}, TSN \cite{tsn}, Motion Vectors \cite{motion-vectors}, 
        and ActionFlow \cite{actionflownet}.
        We show only the accuracy on the motion (temporal) stream for these methods, as we focus only on learning motion representations.
        The TSN method performs the best. Our proposed approach obtains superior performance to ActionFlow \cite{actionflownet}, 
        while having slightly lower performance than Motion Vectors \cite{motion-vectors}, and Two-stream ResNets \cite{spatiotemporal-resnets}, on the temporal stream.
    }
    \label{tab:comparison}
\end{table}

\subsection{Limitations and possible improvements}
Our proposed phase-based motion description shares similar limitations to the classic phase-based approaches, namely dealing with noisy inputs and high-velocity actions. 
Learning an Eulerian transformation from two consecutive frames independently, 
could potentially solve the former problem, while increasing the number of proposed complex layers in the network architecture could help the performance.

Another limitation to keep in mind is the number of distinct overlapping motions patterns per spatial neighborhood. 
Having three or more motion patterns per neighborhood in a video, increases the likelihood of errors in the Eulerian motion representation.
This can happen if the effective size of the receptive field of our perpendicular complex filters in the proposed complex layer is too high. 
A possible improvement to this proposed Eulerian method of learning motion representations for action recognition, is the use of $3$D convolutional filters. 
This may alleviate the problem of achieving limited improvement when stacking Eulerian inputs. 

The project was implemented using TensorFlow. 
The source code for our complex layer can be found at \href{https://github.com/11maxed11/phase-based-action-recognition}{https://github.com/11maxed11/phase-based-action-recognition}.
More details can be found in \cite{omar}.

\section{Conclusions}
\label{sec:conclusion}
We present a new architecture for learning phase-based descriptions from Eulerian inputs, in the context of action recognition. 
The proposed method relies on learning perpendicular complex filters in a ConvNet. 
To help the network learn Gabor-like complex filters we propose a regularization scheme based of frequency analysis, for our learned complex filters. 

Empirical evaluation shows that this architecture delivers an improvement for several Eulerian inputs, while also exceeding the baseline for recognition using a single optical flow input. 
Further improvements of the proposed method are possible, by considering different alternative to boost the performance of the temporal representation by using LSTM layers \cite{lstm1,lstm2}, 
and $3$D convolutional layers \cite{c3d}.

{\small
\bibliographystyle{splncs04}
\bibliography{main}

\begin{thebibliography}{10}
\providecommand{\url}[1]{\texttt{#1}}
\providecommand{\urlprefix}{URL }
\providecommand{\doi}[1]{https://doi.org/#1}

\bibitem{bracewellconvolution}
Bracewell, R.: Convolution" and" two-dimensional convolution." ch. 3 in the
  fourier transform and its applications (1965)

\bibitem{quo-vadis}
Carreira, J., Zisserman, A.: Quo vadis, action recognition? a new model and the
  kinetics dataset. In: CVPR. pp. 4724--4733. IEEE (2017)

\bibitem{lstm1}
Donahue, J., Anne~Hendricks, L., Guadarrama, S., Rohrbach, M., Venugopalan, S.,
  Saenko, K., Darrell, T.: Long-term recurrent convolutional networks for
  visual recognition and description. In: CVPR. pp. 2625--2634 (2015)

\bibitem{spatiotemporal-resnets}
Feichtenhofer, C., Pinz, A., Wildes, R.: Spatiotemporal residual networks for
  video action recognition. In: NIPS. pp. 3468--3476 (2016)

\bibitem{fusion}
Feichtenhofer, C., Pinz, A., Zisserman, A.: Convolutional two-stream network
  fusion for video action recognition. In: CVPR. pp. 1933--1941 (2016)

\bibitem{fleet}
Fleet, D.J., Jepson, A.D.: Computation of component image velocity from local
  phase information. IJCV  \textbf{5}(1),  77--104 (1990)

\bibitem{disparity}
Fleet, D.J., Jepson, A.D., Jenkin, M.R.: Phase-based disparity measurement.
  CVGIP: Image understanding  \textbf{53}(2),  198--210 (1991)

\bibitem{freeman1991design}
Freeman, W.T., Adelson, E.H., et~al.: The design and use of steerable filters.
  TPAMI  \textbf{13}(9),  891--906 (1991)

\bibitem{gautama}
Gautama, T., Van~Hulle, M.M., et~al.: A phase-based approach to the estimation
  of the optical flow field using spatial filtering. TNN  \textbf{13}(5),
  1127--1136 (2002)

\bibitem{omar}
Hommos, O.: Learning Phase-Based Descriptions for Action Recognition. Master's
  thesis, Delft University of Technology (may 2018)

\bibitem{jain201515}
Jain, M., van Gemert, J.C., Snoek, C.G.: What do 15,000 object categories tell
  us about classifying and localizing actions? In: CVPR. pp. 46--55 (2015)

\bibitem{kinetics}
Kay, W., Carreira, J., Simonyan, K., Zhang, B., Hillier, C., Vijayanarasimhan,
  S., Viola, F., Green, T., Back, T., Natsev, P., et~al.: The kinetics human
  action video dataset. CoRR  (2017)

\bibitem{kooij2016depth}
Kooij, J.F., van Gemert, J.C.: Depth-aware motion magnification. In: ECCV. pp.
  467--482 (2016)

\bibitem{phasenet}
Meyer, S., Djelouah, A., McWilliams, B., Sorkine-Hornung, A., Gross, M.,
  Schroers, C.: Phasenet for video frame interpolation. In: CVPR. pp. 498--507
  (2018)

\bibitem{actionflownet}
Ng, J.Y.H., Choi, J., Neumann, J., Davis, L.S.: Actionflownet: Learning motion
  representation for action recognition. CoRR  (2016)

\bibitem{lstm2}
Ng, J.Y.H., Hausknecht, M., Vijayanarasimhan, S., Vinyals, O., Monga, R.,
  Toderici, G.: Beyond short snippets: Deep networks for video classification.
  In: CVPR. pp. 4694--4702. IEEE (2015)

\bibitem{learn-pbmoma}
Oh, T.H., Jaroensri, R., Kim, C., Elgharib, M., Durand, F., Freeman, W.T.,
  Matusik, W.: Learning-based video motion magnification. CoRR  (2018)

\bibitem{pintea2016making}
Pintea, S.L., van Gemert, J.C.: Making a case for learning motion
  representations with phase. In: ECCV workshop. pp. 55--64 (2016)

\bibitem{laura}
Sevilla-Lara, L., Liao, Y., Guney, F., Jampani, V., Geiger, A., Black, M.J.: On
  the integration of optical flow and action recognition. CoRR  (2017)

\bibitem{two-stream}
Simonyan, K., Zisserman, A.: Two-stream convolutional networks for action
  recognition in videos. In: NIPS. pp. 568--576 (2014)

\bibitem{ucf}
Soomro, K., Zamir, A.R., Shah, M.: Ucf101: A dataset of 101 human actions
  classes from videos in the wild. CoRR  (2012)

\bibitem{inception}
Szegedy, C., Liu, W., Jia, Y., Sermanet, P., Reed, S., Anguelov, D., Erhan, D.,
  Vanhoucke, V., Rabinovich, A.: Going deeper with convolutions. In: CVPR (June
  2015)

\bibitem{deepcomplex}
Trabelsi, C., Bilaniuk, O., Zhang, Y., Serdyuk, D., Subramanian, S., Santos,
  J.F., Mehri, S., Rostamzadeh, N., Bengio, Y., Pal, C.J.: Deep complex
  networks. CoRR  (2017)

\bibitem{c3d}
Tran, D., Bourdev, L.D., Fergus, R., Torresani, L., Paluri, M.: C3d: generic
  features for video analysis. CoRR, abs/1412.0767  \textbf{2}(7), ~8 (2014)

\bibitem{varol2018long}
Varol, G., Laptev, I., Schmid, C.: Long-term temporal convolutions for action
  recognition. TPAMI  \textbf{40}(6),  1510--1517 (2018)

\bibitem{pbmoma}
Wadhwa, N., Rubinstein, M., Durand, F., Freeman, W.T.: Phase-based video motion
  processing. TOG  \textbf{32}(4), ~80 (2013)

\bibitem{tsn}
Wang, L., Xiong, Y., Wang, Z., Qiao, Y., Lin, D., Tang, X., Van~Gool, L.:
  Temporal segment networks: Towards good practices for deep action
  recognition. In: ECCV. pp. 20--36. Springer (2016)

\bibitem{rethinking}
Xie, S., Sun, C., Huang, J., Tu, Z., Murphy, K.: Rethinking spatiotemporal
  feature learning for video understanding. CoRR  (2017)

\bibitem{motion-vectors}
Zhang, B., Wang, L., Wang, Z., Qiao, Y., Wang, H.: Real-time action recognition
  with enhanced motion vector cnns. In: CVPR. pp. 2718--2726. IEEE (2016)

\bibitem{zhang2017video}
Zhang, Y., Pintea, S., van Gemert, J.: Video acceleration magnification. In:
  CVPR. IEEE (2017)

\bibitem{hidden-two-stream}
Zhu, Y., Lan, Z., Newsam, S., Hauptmann, A.G.: Hidden two-stream convolutional
  networks for action recognition. CoRR  (2017)

\end{thebibliography}
}

\end{document}